\documentclass[letterpaper]{article} 
\usepackage{apreprint}  
\pdfoutput=1
\usepackage{times}  
\usepackage{helvet} 
\usepackage{courier}  
\usepackage[hyphens]{url}  
\usepackage{graphicx} 
\usepackage{appendix} 
\usepackage{import} 
\usepackage{subfiles} 
\usepackage{graphicx}  
\usepackage{array}
\usepackage[utf8]{inputenc} 
\usepackage[T1]{fontenc}    
\usepackage{booktabs}       
\usepackage{amsfonts}       
\usepackage{nicefrac}       
\usepackage{microtype}      

\usepackage{graphics}
\usepackage{graphicx}
\usepackage{algorithm}
\usepackage{algorithmicx}
\usepackage[noend]{algpseudocode}
\usepackage{amssymb}
\usepackage{amsmath }
\usepackage{subfigure}

\algnewcommand{\LineComment}[1]{\State \(//\) #1}

\newlength\myindent
\begin{document}
\title{The Neural State Pushdown Automata}
\author{
  Ankur Mali\textsuperscript{\rm 1} \\
  \texttt{aam35@psu.edu,} \\
   \And
 Alexander Ororbia\textsuperscript{\rm 2} \\
  \texttt{ago@cs.rit.edu} \\
  \\
    \textsuperscript{\rm 1}Penn State University\\ State College, PA 16801\\
    \textsuperscript{\rm 2}Rochester Institute of Technology\\
    Rochester, NY 14623\\
  \And
 C. Lee Giles \textsuperscript{\rm 1}\\
  \texttt{clg20@psu.edu } \\
}

\maketitle 
\begin{abstract} 
In order to learn complex grammars, recurrent neural networks (RNNs) require sufficient computational resources to ensure correct grammar recognition. A widely-used approach to expand model capacity would be to couple an RNN to an external memory stack. Here, we introduce a ``neural state'' pushdown automaton (NSPDA), which consists of a digital stack, instead of an analog one, that is coupled to a neural network state machine.  We empirically show its effectiveness in recognizing various context-free grammars (CFGs). First, we develop the underlying mechanics of the proposed higher order recurrent network and its manipulation of a stack as well as how to stably program its underlying pushdown automaton (PDA) to achieve desired finite-state network dynamics. 
Next, we introduce a noise regularization scheme for higher-order (tensor) networks, to our knowledge the first of its kind, and design an algorithm for improved incremental learning.
Finally, we design a method for inserting grammar rules into a NSPDA and empirically show that this prior knowledge improves its training convergence time by an order of magnitude and, in some cases, leads to better generalization. The NSPDA is also compared to a classical analog stack neural network pushdown automaton (NNPDA) as well as a wide array of first and second-order RNNs with and without external memory, trained using different learning algorithms. 
Our results show that, for Dyck(2) languages, prior rule-based knowledge is critical for optimization convergence and for ensuring generalization to longer sequences at test time.  We observe that many RNNs with and without memory, but no prior knowledge, fail to converge and generalize poorly on CFGs.
 
\end{abstract}
\section{Introduction}
\label{sec:intro}
\noindent
Despite their success, artificial neural networks (ANNs), especially recurrent neural networks (RNNs), have repeatedly been shown to struggle with generalizing in a sophisticated, systematic manner, often uncovering misleading statistical associations instead of true casual relations. Verifying what is learned by these black-box models remains an open challenge, centering around one central issue -- the lack of interpretability and modularity. The fact that successful ANN optimization depends heavily on large quantities of data only serves to further worsen the problem.

One research direction towards developing more interpretable ANNs focuses on rule extraction from and assimilation of rules into RNNs \cite{angluin1983inductive,fu1977}. To solve difficult grammatical inference problems, various types of specialized RNNs have been designed \cite{lstmcfg,boden2000context,tabor2000fractal,wiles1995learning,sennhauser2018evaluating,nam2019number}
However, it has been shown that RNNs augmented with external memory structures, such as the neural network pushdown automaton (NNPDA), are more powerful than RNNs without, both historically \cite{giles1992learning,pollack1990recursive,zeng1994discrete} and recently, using differentiable memory \cite{joulin2015inferring,grefenstette2015learning,graves2014neural,kurach2015neural,zeng1994discrete,hao2018context,yogatama2018memory,graves2016hybrid}.
Yet most of these models often lack interpretability and how they learn any given grammar is still debatable. In the past, rule integration methods have been proposed to tackle the interpretability issue \cite{giles1992learning,omlin1996constructing} and offer a promising path towards the design of ANNs with an underlying knowledge structure that is bit more understandable and transparent. 
However, to the best of our knowledge, there exists no method for inserting rules into the states of the far more powerful class of higher order, memory-augmented RNNs.

In working towards interpretable, memory-based neural models, in this work, our contributions are the following: 
\begin{itemize}
\item We propose the neural state pushdown automaton and its incremental training method, which exploits the concept of iterative refinement 
\item We develop a novel regularization method that empirically yields better generalization in complex, memory-based RNNs. To our knowledge, we are the first to propose a weight regularizer that works with higher-order RNNs. 
\item We propose a method for programming states into a neural state machine with binary second and third-order weights .
\item We develop a method for inserting rules into stack-based recurrent networks.
\item We compare our model with the NNPDA and other RNNs, trained using different learning algorithms. 
\end{itemize}

\section{Motivation \& Related Work}
\label{sec:motivation}


Research related to integrating knowledge into ANNs has existed for quite some time, such as through the design of state machines \cite{tivno1998finite,omlin1996constructing}. Recent efforts in the domain of natural language processing have shown the effectiveness of using state machines for tasks such as visual question answering, which allow an agent to directly use higher-level semantic concepts to represent visual and linguistic modalities \cite{manning2019nsm}. With respect to rule-insertion itself, there exists a great deal of work showcasing its effectiveness when used with ANNs\cite{abuMostafa1990hints} as well as with RNNs \cite{giles1992learning,omlin1996constructing}. 
Notably, \cite{omlin1996constructing} showed how deterministic finite automaton rules could be encoded into second order RNNs.

One important, classical model that we draw inspiration from is the neural network pushdown automaton (NNPDA) \cite{nndpa1998sun}. The structure of our proposed model is similar to the NNPDA, but, as we will discuss, the major difference is that the model works with a digital stack as opposed to an analog one. 
Interestingly enough, prior work has also shown how to ``hints'' into the NNPDA, where knowledge of ``dead states'' can be used to guide its learning process \cite{nndpa1998sun}. In the spirit of this hint-based methodology, we will develop a method for encoding useful rules related to target CFGs into our neural state pushdown automaton (NSPDA). This, to our knowledge, is the first approach of its kind, since no rule methodology has been previously proposed for complex state-based models. Creating such a procedure allows us to both exploit the far greater representational capabilities of memory-augmented RNNs while offering an intuitive way for understanding the knowledge contained and acquired by RNNs.

In this work, we will focus on RNNs that control a discrete stack, particularly our proposed NSPDA. We will empirically determine if the inductive biases we encode into its synaptic weights speed up the parameter optimization process and, furthermore, improve model generalization over longer sequences at test time. 
Furthermore, the results of our experiments, which compare a wide variety of RNNs (of varying order, with and without memory), will strongly contradict the claim presented in recent work \cite{gru2019pda}, which specifically claims that first order RNNs, like the popular gated recurrent unit RNN \cite{chung2014empirical}, are as powerful as a PDA. In essence, our work demonstrates that for an RNN to recognize a complex CFG, it will, at least, require external memory. Our results also demonstrate the value of encoding even partial PDA information which positively impacts convergence time and model generalization.

\section{The Neural State Pushdown Automaton}
\label{sec:nspda}

\subsection{Neural Architecture}
\label{sec:arch}
\begin{figure}
    \centering
    \includegraphics[width=115mm]{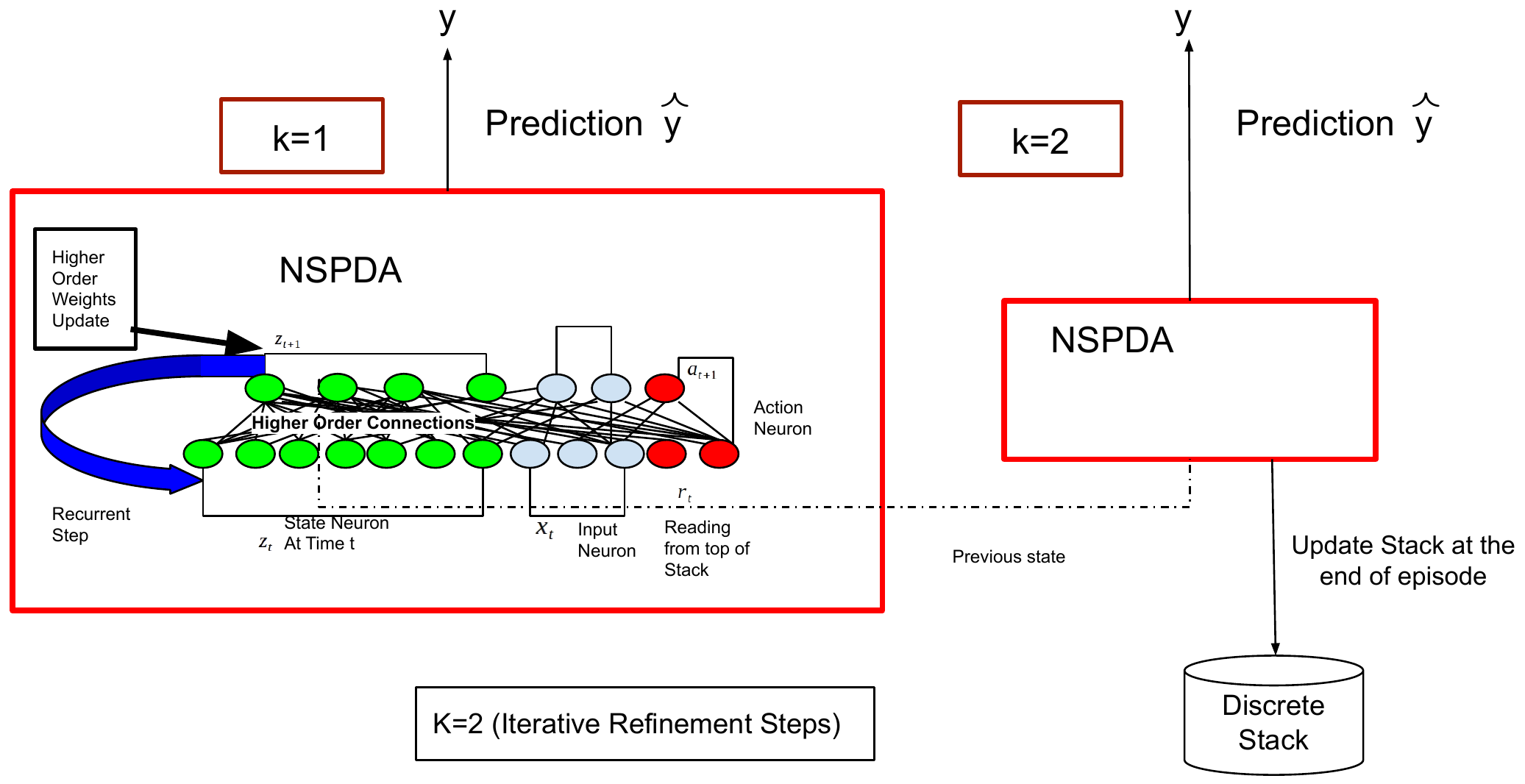}
    \caption{The NSPDA shown making predictions over $K=2$ steps of iterative refinement.} 
    \label{fig:model_diagram}
\end{figure}

The model we propose, the NSPDA with iterative refinement is shown in figure \ref{fig:model_diagram}. The NSPDA consists of fully connected recurrent neurons which we will label as \emph{state neurons}, primarily to distinguish them from the neurons that function as output neurons. Introducing the concept of state neurons is important when considering the notion of higher-order networks, i.e., second or third order RNNs, which allows us to map state representations directly to outputs. In this model, at each time step $t$, the state neuron receives signals from the input neurons, its previous state, and the stack-read neurons. The input neurons process a string, one character at a time, while non-recurrent neurons, also labeled as ``action neurons'', represent an operation to be performed on a stack data structure, i.e., Push/Pop/No-op. The action neurons are also designated as the controller which can either be recurrent or linear (recurrent controllers usually perform better in practice, so we focus on these in this paper). Furthermore, ``read'' neurons are used to keep track of the symbols present at the top of the stack. 

To make concrete the above high-level description, consider a single hidden-layer NSPDA. A full symbol sequence sample $(y,\mathbf{X})$ is defined as $\mathbf{X} = \{\mathbf{x}_1,\mathbf{x}_2,\cdots,\mathbf{x}_T\}$ where the binary label $y$ indicates whether the sequence is valid ($1$) or not ($0$). When processing a (binary) symbol/token $\mathbf{x}_t \in \{0,1\}^{L \times 1}$ at the discrete time step $t$, the NSPDA is engaged with computing a new state variable vector $\mathbf{z}_t \in \mathbb{R}^{J \times 1}$, where $L$ is the total number of input/sensory neurons (or dimensionality of the input space, sometimes classically refered to as alphabet size) and $J$ is the total number of state neurons. The action neuron vector is defined as $\mathbf{a} \in \mathbb{R}^{L \times 1}$ and the read neuron vector is defined as $\mathbf{r} \in \mathbb{R}^{L \times 1}$, i.e., the action and read spaces are of the same dimensionality of the input or $|\mathbf{x}| = |\mathbf{r}| = |\mathbf{a}|$.
Taken together, the above sets of input, state, and read neurons represent a full NSPDA model with parameters $\Theta = \{W^s, W^a, W^o\}$. Crucially, $W^s$ and $W^a$ are both 4-dimensional (4D) synpatic weight tensor, i.e., the binary ``to-state'' tensor $W^s \in \{0,1\}^{J \times L \times L \times L}$ and the 4D tenary to-action tensor $W^a \in \{-1,0,1\}^{J \times L \times L \times L}$ (note that: $-1$ is ``pop'', $0$ is ``no-op'', and $1$ is ``push''). At $t$, inference (for a third order NSPDA) is conducted as follows:
\begin{align}
    z_{t+1}^i &= g(\Sigma_{j,k,l}W_{ijkl}^s(z_t^j,r_t^k,x_t^l)+b_s^i) \label{eqn:state}
    \\
    a_{t+1}^i &= f(\Sigma_{j,k,l}W_{ijkl}^a(z_t^j,r_t^k,x_t^l)+b_a^i) \label{eqn:action}
    \\
    r^i_{t+1} &=
    \left\{ 
    \begin{array}{rl}
      \alpha_1 & \text{if }a^i_t = 0 \\ 
      \alpha_2 & \text{if }a^i_t = 1 \\ 
      \alpha_3 & \text{if }a^i_t = -1   
    \end{array}
    \right.\ \label{eqn:read}
\end{align}
where $\alpha_1 \sim U(0.0001,0.008)$, $\alpha_2 \sim U(0.901,0.992)$, and $\alpha_3 \sim U(0.025,0.110)$, are threshold values that determine what the next state of the discrete read unit $r^i_t$ will be (sampled uniformly from a special interval to create continuous value for backprop to work with). 
Note that $\mathbf{z}_{t+1}$ is the next hidden state, $\mathbf{a}_{t+1}$ is the next stack action, and $\mathbf{r}_{t+1}$ is the next value of the neuron that reads the content at the top of the stack. 
$g(v)$ and $f(v)$ are non-linear activation functions, specifically, quantized sigmoidal functions, defined as:
\begin{align}
    \hat{g}(v) &= \frac{1}{(1 + e^{-v})}, \quad \hat{f}(v) = 2g(v) - 1 \\
    g(v) &= 
    \left\{ 
    \begin{array}{rl}
      1 & \text{if }\hat{g}(v) > 0.5 \\ 
      0 & \text{otherwise.}
    \end{array}
    \right.\ \\
    f(v) &= 
    \left\{ 
    \begin{array}{rl}
      1 & \text{if }\hat{f}(v) > 0.13 \\ 
      0 & \text{if }-0.09 \leq \hat{f}(v) \leq 0.13 \\ 
      -1 & \text{otherwise.}
    \end{array}
    \right.\
\end{align}
As the NSPDA processes a string, a prediction $\hat{y}_t$ of its validity is made at each step. Specifically, the output weights $W^o \in \mathbb{R}^{J \times 1}$ (and bias scalar $b^o$) are used to map the state vector $\mathbf{z}_t$ to the output space. The output model is defined as $\hat{y}_t = \sigma( W^o \cdot \mathbf{z}_t + b^o ) $, where $\sigma(v)$ is the logistic link function.

The actual external stack itself is manipulated by discrete-valued action neurons that trigger a discrete push or pop action (as given by Equation \ref{eqn:action}). Take, for example, a 2-letter alphabet, i.e., $\{a,b\}$. The dimensions of the action and read spaces would then, in this case, be $|\mathbf{a}| = |\mathbf{r}| = 2$.
When using a digital stack, the following actions can be taken:
\begin{itemize}
\item \textbf{PUSH:}
This means that the current input is pushed to the top of the stack. Example: To push the symbol ``a'', use $\mathbf{a}_t = <1, 0 >$ and $\mathbf{r}_t = <0.955, 0.008 >$.

\item \textbf{POP:} 
This means that the element is removed from the top of the stack. Example: To remove the symbol ``b'', use $\mathbf{a}_t = <0, -1 >$ and $\mathbf{r}_t = <0.008, 0.065 >$.

\item \textbf{NO-OP:} This simply means ``no operation, or, in other words, nothing is to be done with the stack. Example: use $\mathbf{a}_t = <0, 0 >$ and $\mathbf{r}_t = <0.008, 0.008 >$.

\end{itemize}
In the case of the vector $\mathbf{r}_t$, we are reading the symbol currently located at the top of the stack (at each time step) (corresponding read vectors are shown above in the action vector examples0.
Our goal is to make sure the RNNs choose the correct action during training and yet still maintain stable binary read states $\mathbf{z}_t$. 

\subsection{Learning and Optimization}
\label{sec:opt}

First, we define the loss function used to both measure the performance of the network as well as optimize its parameters. Classically, state neural models such as the NNPDA exclusively made use of a binary loss function that only considered if a string was valid or invalid \cite{das1993using}. Furthermore, these models only made a prediction/classification at the very end of the sequence. In contrast, the NSPDA is an iterative, step-by-step predictive model. Thus, we consider using a sequence loss based on binary cross entropy.\footnote{In preliminary experiments, models using a squared error loss, with and without regularization penalties, had great difficulty in converging. We found using cross entropy was far more effective. }. The instantaneous loss, for a single sequence $(y,\mathbf{X}$), is:
\begin{align}
    \mathcal{L}(y,\mathbf{X},\Theta) = \sum^T_{t=1} -y \log(\hat{y}_t)) - (1 - y)\log(1 - \hat{y}_t)  \mbox{.}
    \label{eqn:loss}
\end{align}
where $\hat{y}_t$ is the $t$-th prediction/output from the final state neuron. Note that $y$ is copied each step in time, which injects an extra error signal throughout the sequence length, improving the optimization process (as opposed to relying on only a single output error signal to be effectively propagated backwards through the underlying computation graph).

To compute updates for the NSPDA's parameters, we employed several gradient-based approaches, including the popular and common back-propagation through time (BPTT) procedure as well as online algorithms such as real-time recurrent learning (RTRL) \cite{williams1989rtrl} and unbiased online recurrent optimization (UORO) \cite{tallec2017uoro}. In short, all of these algorithms compute gradients of the loss function (Equation \ref{eqn:loss}) with respect to NSPDA weights. The primary difference between the algorithms is that BPTT is based on reverse-mode differentiation routine while RTRL is based on forward-mode differentiation (and UORO is a faster, higher variance approximation of RTRL). In further detail, we describe UORO and RTRL in the appendix. While UORO and RTRL are not commonly used to train modern-day RNNs, they offer faster ways to train them without requiring graph unfolding. Thus, we compare the results of using each in our experiments.

\subsubsection{Iterative Refinement}
\label{sec:iter_refine}
One important element we introduced into the training protocol of the NSPDA is that of iterative refinement, an algorithm proposed in the signal processing literature for incorporating partial iterative inference into a next-step predictive RNN \cite{ororbia2019iterdecode}. At a high-level, this means that, during training, at step $t$, the NSPDA is forced to predict the same target ($\mathbf{y}_t$) $K$ times (except for the state transitions that are provided as ``hints'', of which we will describe in a later section). Crucially, the state vector is still carried over these $K$ steps, meaning the recurrent synapses relating the state of the model at time $t$ to $t+1$ .To adapt iterative refinement to a next-step sequence model like the NSPDA, iterative refinement can cleanly introduced by manipulating the sequence loss of Equation \ref{eqn:loss} as follows:
\begin{align}
    \mathcal{D}(\hat{y},y) &= -y \log(\hat{y})) - (1 - y)\log(1 - \hat{y}) \\
    \mathcal{L}(y,\mathbf{X},\mathbf{S},\Theta) &= \sum_t \sum^{K = \mathbf{S}(t)}_{k=1} \mathcal{D}(\hat{y}_{t,k},y)
    \label{eqn:if_loss}
\end{align}
noting that we have introduced the variable $\mathbf{S}$ to augment the sample $(y,\mathbf{X})$. $\mathbf{S}$ is an integer sequence computed as follows: $\mathbf{S} = K (1 - \mathbf{H}) + \mathbf{H}$ where $\mathbf{H}$ is a binary ``hint'' vector (automatically generated) of the form $\mathbf{H} = \{h_0,h_1,\cdots,h_T \}$ ($h_t = 1$ signals a hint is used, while $h_t = =0$ is ``no hint''). Empirically, we found $K=4$ worked well.
In \cite{ororbia2019iterdecode}, using an RNN's recurrent weights as a lateral processing mechanism \cite{ororbia2019lifelong} was related to an RNN acting as a deep feedforward network with tied weights across $K$ hidden layers (a ``prediction episode''). This means that additional nonlinearity (via depth) is being efficiently exploited without incurring the memory cost of storing extra weights. We found that iterative refinement introduces greater stability into learning process primarily when gradient noise is used. Note that, even in this case, while we work with full precision weights for gradient computation, before evaluation is conducted, the weights are converted to discrete values. 

\subsubsection{Two Stage Incremental learning}
\label{sec:incremental}
Incremental learning, or, in other words, training procedures that sort data samples based on their inherent difficulty and progressively present them to a neural agent progressively, has been shown to quite effective when training RNNs on input data that is known to have some structure \cite{elman1993learning,das1993using}. 
Based on this prior finding, we developed a two-stage incremental learning approach for improving a higher-order RNN's ability to generalize to longer sequences. Formally, Algorithm \ref{alg:increment_learning} depicts the overall process. We found that using a stochastic learning rate \cite{ororbia2019iterdecode} worked better in the first stage while a fixed learning rate combined with stochastic noise process applied to the weights (similar to gradient noise) worked better during second stage.
\begin{algorithm}
\caption{Two Stage Incremental Learning}
\label{alg:increment_learning}
\begin{algorithmic}
    \State {\bfseries Input:} $\Theta$ (model weights), training set $\mathcal{D}$, validation set $\mathcal{V}$, $N_{Tr}$ (midpoint length threshold), $\lambda$ (learning rate)
    \LineComment --------------------------- Stage \#1 ---------------------------
    \State $N_{max} = maxLen(\mathcal{D}$) \Comment Calculate longest string length
    \LineComment Sequential Curriculum Update Phase
    \State $\mathcal{D}_l = \emptyset$
    \For{$N_l=1$ to $N_{Tr}$}
        \State $\mathcal{D}_l$ = Extract from $\mathcal{D}$ all strings lengths $\le N_l$
        \State \emph{TRAIN}(Model($\Theta$), $\lambda$, $\mathcal{D}_l$) \Comment Single pass through $\mathcal{D}_l$
    \EndFor
    \LineComment Random Curriculum Phase 
    \While {Model($\Theta$) not converged on $\mathcal{V}$ or $e_n < 200$}
        \State \emph{TRAIN}(Model($\Theta$), $\lambda$, $\mathcal{D}_l$),\quad $e_n = e_n + 1$
    \EndWhile

    \LineComment --------------------------- Stage \#2 ---------------------------
    \LineComment Sequential Curriculum Update Phase
    \State $\mathcal{D}_l = \emptyset$
    \For{$N_l=1$ to $N_{max}$} 
        \State $\mathcal{D}_l$ = Extract from $\mathcal{D}$ all strings lengths $\le N_l$
        \State \emph{TRAIN}(Model($\Theta$), $\lambda$, $\mathcal{D}_l$) \Comment Single pass through $\mathcal{D}_l$
    \EndFor
    \LineComment Random Curriculum Phrase
    \While {Model($\Theta$) not converged on $\mathcal{V}$ or $e_n < 350$}
        \State \emph{TRAIN}(Model($\Theta$), $\lambda$, $\mathcal{D}_l$),\quad $e_n = e_n + 1$
    \EndWhile
    \State \Return $\Theta$ \Comment Return final trained model weights
   \end{algorithmic}
\end{algorithm}
As we will see later experimentally, whenever the data has some exploitable structure that allows for an automatic sorting of samples by increasing complexity, incremental learning is highly effective in training higher-order RNNs. In the case of CFGs, we can sort samples based on string length and progressively build a model that can learn to generalize to increasingly longer string sequences. Algorithm \ref{alg:increment_learning} depicts the full process (note that we set $N_T = 14$ in this paper and $e_n$ is a variable that marks the number of epochs so far).

\subsubsection{Regularizing Higher Order RNNs:}
\label{sec:regularization}
When training any RNN for long periods of time, the model tends to memorize the input training data which damages its ability to generalize to unseen sequence data, i.e., overfitting. Higher order RNNs are also susceptible to overfitting given their  high-capacity and complexity, and yet, no regularization has ever been proposed to help these kinds of RNNs to combat overfitting. In this work we extend an adaptive (layer-dependent) noise scheme that was originally proposed for training neurobiologically-plausible ANNs \cite{ororbia2019biologically}, which showed strong positive results for simple feedforward classification tasks, to RNNs. Notably, our noise-based regularizer applies to higher-dimensional tensors, which are fundamental to implementing any $n$-th order RNN.  
\begin{algorithm}
\caption{Adaptive Noise Regularizer}
\label{alg:dropout_NSPDA}
\begin{algorithmic}
    \State {\bfseries Input:} Tensor $W \in \mathbb{R}^{A \times B \times C \times D}$, e.g., $W^s$ or $W^a$
    \LineComment $N_p$=Percentage of Noise, ``$\cdots$'' means $k=k+1$
    \Function{CreatePartitions}{W, K,$N_p$} \Comment{Partition sub-routine for noise regularization function}
        \LineComment $len(W) = A * B$ (calculate length by multiplying 1st two tensor dimensions) 
        \LineComment $s(P_i,N_p)$ randomly selects $N_p$ matrices in $P_i$
        \LineComment Divide W into $3$ partitions $\{P_1,P_2,P_3\}$ 
        \LineComment $P_i = \{M_1,M_2,\cdots,M_{len(W)} \}$, $M_i \in \mathbb{R}^{C \times D}$ 
        \State \textbf{if} $len(W)$ is odd
        \State \quad $P_1 = W[k=1,\cdots,K/3]$
        \State \quad $P_2 = W[k=k/3+1,\cdots,2K/3]$
        \State \quad $P_3 = W[k=(2k/3)+1,\cdots,K]$
        \State \textbf{else} $len(W)$ is even
        \State \quad $P_1 = W[k=1,\cdots,(K-1)/3]$
        \State \quad $P_2 = W[k=(k-1)/3+1,\cdots,(2K-1)/3]$
        \State \quad $P_3 = W[k=(2k-1)/3+1,\cdots,K]$
        
        \LineComment Create set $Q$ of $N_p$ random matrices from each ${P_{i}}$
        \State $Q = \{s(P_1,N_p),s(P_2,N_p),s(P_3,N_p)\}$
        \State \textbf{Return} $Q$
    \EndFunction

    \Function{Adaptive Noise}{$Q$}
        \State $\widehat{p} \sim \mathcal{N}(\mu=0,\sigma=1)$ \Comment Draw Gaussian scalar sample
        \For{each $M$ in  $Q$ } \Comment for each matrix in $Q$
            \State $M = (\widehat{p} * \beta) M$, $\widehat{p}= \widehat{p} / 2$
        \EndFor
        \LineComment Remap matrices $M$ in $Q$ to tensor shaped like $W$
        \State $W \leftarrow remap($Q$)$
        \State \textbf{Return} $W$
        
        \LineComment Use updated weight matrix for gradient computation
        
  \EndFunction
   
   \end{algorithmic}
\end{algorithm}
We are also motivated by the fact that injecting noise to gradients can encourage exploration of an RNN's error optimization landscape \cite{Goodfellow16} in one of two ways: 1) at the input, i.e., data augmentation \cite{Goodfellow16}, or 2) at the recurrence \cite{krueger2016zoneout}.
Our regularizer falls under the second case.\footnote{We implemented a data augmentation approach but found it yielded poor results when learning context-free grammars.}

The key details of our noise-based regularizer are depicted in Algorithm \ref{alg:dropout_NSPDA}. Based on preliminary experiments, we found that a noise level less than 30\% and more than 8\% helps the network to converge faster and, more importantly, generalize better on unseen sequences, longer that than those found in the training set. Experimentally, later we will see that this regularizer improves generalization even when prior knowledge is not integrated into the RNN. 

\section{Integrating Prior Knowledge}
\label{sec:priors}

\subsection{Programming and Inserting Rules}
\label{sec:programming_rules}
We start by defining the data generating process that any RNN is to learn from, i.e., a PDA that generates a set of positive and negative strings.  
Formally, the $M$-state PDA is defined as a 7-tuple $(Q,\Sigma,\Gamma, \delta,{q^0},\bot,F)$ where:
\begin{itemize}
    \item $\Sigma = \{a^1,\cdots,a^l,\cdots,a^L\}$ is the input alphabet
    \item $Q = \{s^1,\cdots,s^m,\cdots,s^M\}$ is the finite set of states
    \item  $\Gamma$ is known as stack alphabet (a finite set of tokens)
    \item ${q^0}$ is the start state
    \item  $\bot$ is the initial stack symbol
    \item $F \subseteq Q$ is the set of accepting states
    \item $\delta \subseteq Q \times (\Sigma \cup )| \times \Gamma \rightarrow Q \times \Gamma^*)$ is the state transition.
\end{itemize}
To insert rules related to known state transitions into the ($N$-state) NSPDA, one needs to program its recurrent weights (which could be second or third order).
Since the number of states in PDA is not known before hand, we assume that $J > M$ and that the network has enough capacity to learn an unknown context-free grammar.

In order to program and insert rules, we propose adapting methodology originally developed for second-order RNNs and deterministic finite state automata (DFA) \cite{omlin1996constructing} to the case of PDA-based RNNs. Specifically, we will exploit the similarity between the state transitions of the target PDA and the underlying dynamics of a stack-driven RNN.
Consider a known transition $\delta (s^j,a^l,\mathbf{T_s}) = (s^i,\gamma)$, where $\mathbf{T_s}$ is the top of the stack and $\gamma$ is the sequence of symbols replacing $\mathbf{T_s}$.
We then identify PDA states $s^j$ and $s^i$, which correspond to state neurons $z^j$ and $z^i$, respectively. Recall that each symbol has specific stack operations associated with it, which provide prior knowledge as to when to push and when to pop from the stack. 
It is desirable that the state neuron $z^i$ has a high output close to $1$ and $z^j$ has a low output close to $0$ after reading an input symbol $a^l$ using input neuron $x^m$ and the top of the stack $\mathbf{T_s}$ using read neuron $r^l$ (remember that a read depends on an action neuron, as depicted in model Equation \ref{eqn:read}). This condition can be achieved by doing the following: 1) set the (third order) weights $W_{ijkl}^s$ to a large positive value, which helps to ensure that the state neuron $z_i^{t+1}$ at the next time step $t+1$ will be high (and since $\hat{g}(v)$ is sigmoidal, this tends towards $1$), 
and 2) set $W_{jjkl}^s$ to a large negative value, which would make the output of the state neuron $z_j^{t+1}$ low (tending $\hat{g}(v)$ towards $0$).

The next item to consider are the (ternary) action weights stored in  $W_{ijkl}^a$, which drive the action neurons that yield the stack operations (recall that [-1,0,1] maps to [pop,no-op,push]).
First, we must assume that the total contribution of the weighted output of all state neurons can be neglected -- this can be achieved by setting all other state neurons to the lowest value. In addition, we assume that each state neuron can only be assigned to one known state of the PDA.
If we have prior knowledge of accepting and non-accepting states related to a particular neuron, we may then bias its output $z^i_{t+1}$. We start from $i=1$ (the leftmost neuron in the vector $\mathbf{z}_t$) and work towards $i=J$, programming each one by one. Armed with these assumptions, we can then stably encode rules into the NSPDA by programming the weight $W_{ijkl}^s$ to be large positive value if the PDA's state $s^i$ is an accepting state. Otherwise, we set  $W_{ijkl}^s$ to be a large negative value if the state is non-accepting. If no such knowledge of the PDA is available, $W_{ijkl}^s$ remains unchanged.

Though described for a third order NSPDA, the above approach for programming weights also applies to a second order model as well. In a lower order NSPDA, with 3D weight tensors $W_{ijk}^s$ and $W_{ijk}^a$, state updates and transitions are conducted by concatenating a read neuron $r^i_t$ with an input neuron $x^k_t$ to create a single vector. However, when programming a second order model, we are now working with a DFA \cite{omlin1996constructing} instead of a PDA, which limits the capabilities of the NSPDA (as well as restricts its capacity) since we do not possess any knowledge about what to push or pop. However, when combined with our proposed learning procedure that incorporates iterative refinement, we believe that the second order NSPDA can still learn what action to perform. However, the issue of dimensionality arises -- the state space of a lower order model is very large when compared to that of a third order NSPDA. 
In the case of a PDA-based model, pushing multiple symbols might lead to reaching same accepting state, however, in case of a DFA-based model (the second order NSPDA), we create separate sets of accepting states for each symbol. We found that this splitting mechanism was crucial in getting our network to work perfectly with a digital stack.

While the above rule insertion scheme seems simple enough, determining the actual values for the weights that are to be programmed can be quite problematic. In the case of third order synaptic connections (with binary weights), with just $4$ neurons, there are  $2^{256}$ different combinations, which would quickly render our method impractical and near useless.
However, we can sidestep this computational infeasibility by making use of ``hints'' \cite{omlin1996constructing} within the framework of ``orthogonal state encoding''. By assuming that the PDA starts generating a valid grammar at its initial state, we can then randomly choose a single state and make the output of one state neuron equal to $1$. The outputs of all the other neurons are set to be equal to $0$. Following this, we set the values of weights (according to known state transitions) according to the approach described above. Notably, these weights, though initially programmed, are still adaptable, making them amenable to tuning to a target grammar underlying a data sample.
Programming the weights of second or third order networks jointly impacts the behavior of the state neurons $\mathbf{z}_t$, the read neurons $\mathbf{r}_t$ and the input neurons $\mathbf{x}_t$. Following the scheme we described above yields sparse NSPDA representations of PDA states.

It is difficult to program an NSPDA with a minimal number of states, despite the fact that we have a theoretical guarantee that the third order model is equivalent to PDA dynamics \cite{nndpa1998sun}.

We will observe in our results, the proposed methodology significantly reduces the NSPDA's convergence time during optimization (leading to roughly comparable training time characteristic of first order RNNs), which is particularly important given the fact that its inference process entails 4D tensor products (which are far more expensive than the matrix computations of modern-day RNNs).

\begin{table*}[!t]

\centering
\begin{tabular}{l|r|r||r|r||r|r||r|r}
\multicolumn{1}{l}{}&\multicolumn{2}{|c}{\begin{tabular}[x]{@{}c@{}}\textbf{$Palindrome$}\\\end{tabular}}&\multicolumn{2}{|c}{\begin{tabular}[x]{@{}c@{}}\textbf{$a^nb^n$}\\\end{tabular}}&\multicolumn{2}{|c}{\begin{tabular}[x]{@{}c@{}}\textbf{$a^nb^ncb^ma^m$}\\\end{tabular}}&\multicolumn{2}{|c}{\begin{tabular}[x]{@{}c@{}}\textbf{$a^{n+m}b^nc^m$}\\\end{tabular}}\tabularnewline
\multicolumn{1}{l}{\textbf{Rule Method}}&\multicolumn{1}{|c}{\begin{tabular}[x]{@{}c@{}}\textbf{W1}\\\end{tabular}}&\multicolumn{1}{|c}{\begin{tabular}[x]{@{}c@{}}\textbf{W2}\\\end{tabular}}&\multicolumn{1}{|c}{\begin{tabular}[x]{@{}c@{}}\textbf{W1}\\\end{tabular}}&\multicolumn{1}{|c}{\begin{tabular}[x]{@{}c@{}}\textbf{W2}\\\end{tabular}}&\multicolumn{1}{|c}{\begin{tabular}[x]{@{}c@{}}\textbf{W1}\\\end{tabular}}&\multicolumn{1}{|c}{\begin{tabular}[x]{@{}c@{}}\textbf{W2}\\\end{tabular}}&\multicolumn{1}{|c}{\begin{tabular}[x]{@{}c@{}}\textbf{W1}\\\end{tabular}}&\multicolumn{1}{|c}{\begin{tabular}[x]{@{}c@{}}\textbf{W2}\\\end{tabular}}\tabularnewline
\hline
\textit{NNPDA w/o hints} & $100$ & $81$ & $280$ & $215$ & $NA$ & $NA$ & $NA$ & $NA$ \tabularnewline
\textit{NNPDA w/ dead neuron hints} & $92$ & $83$ & $212$ & $192$ & $485$ & $145$ & $250$ & $195$ \tabularnewline
\textit{NSPDA w/o hints} & $91$ & $79$ & $221$ & $192$ & $488$ & $159$ & $339$ & $293$ \tabularnewline
\textit{NSPDA w/ Hint \#1} & $80$ & $75$ & $190$ & $170$ & $410$ & $140$ & $240$ & $160$ \tabularnewline
\textit{NSPDA w/ Hint \#2} & $\textbf{70}$ & $\textbf{72}$ & $\textbf{150}$ & $\textbf{138}$ & $\textbf{389}$ & $\textbf{134}$ & $\textbf{222}$ & $\textbf{148}$ \tabularnewline
\hline
\end{tabular}

\vspace{-0.1cm}
\caption{Comparison between NSPDAs trained w/ and w/o hints using either $2$nd order weights (W1) or $3$rd order weights (W2).}
\label{hint_results}

\centering

\begin{tabular}{l|r|r||r|r||r|r||r|r}
\multicolumn{1}{l}{}&\multicolumn{2}{|c}{\begin{tabular}[x]{@{}c@{}}\textbf{$Palindrome$}\\\end{tabular}}&\multicolumn{2}{|c}{\begin{tabular}[x]{@{}c@{}}\textbf{$a^nb^n$}\\\end{tabular}}&\multicolumn{2}{|c}{\begin{tabular}[x]{@{}c@{}}\textbf{$a^nb^ncb^ma^m$}\\\end{tabular}}&\multicolumn{2}{|c}{\begin{tabular}[x]{@{}c@{}}\textbf{$a^{n+m}b^nc^m$}\\\end{tabular}}\tabularnewline
\multicolumn{1}{l}{\textbf{Train Method}}&\multicolumn{1}{|c}{\begin{tabular}[x]{@{}c@{}}\textbf{M1}\\\end{tabular}}&\multicolumn{1}{|c}{\begin{tabular}[x]{@{}c@{}}\textbf{M2}\\\end{tabular}}&\multicolumn{1}{|c}{\begin{tabular}[x]{@{}c@{}}\textbf{M1}\\\end{tabular}}&\multicolumn{1}{|c}{\begin{tabular}[x]{@{}c@{}}\textbf{M2}\\\end{tabular}}&\multicolumn{1}{|c}{\begin{tabular}[x]{@{}c@{}}\textbf{M1}\\\end{tabular}}&\multicolumn{1}{|c}{\begin{tabular}[x]{@{}c@{}}\textbf{M2}\\\end{tabular}}&\multicolumn{1}{|c}{\begin{tabular}[x]{@{}c@{}}\textbf{M1}\\\end{tabular}}&\multicolumn{1}{|c}{\begin{tabular}[x]{@{}c@{}}\textbf{M2}\\\end{tabular}}\tabularnewline
\hline
\textit{Standard} & $5699$ & $5912$ & $>200000$ & $>210000$ & $>240000$ & $>233000$ & $>320000$ & $>315000$ \tabularnewline
\textit{IL} & $2678$ & $2552$ & $108200$ & $104556$ & $192001$ & $192551$ & $222171$ & $222144$ \tabularnewline
\textit{2-IL} \textbf{(ours)} & $\textbf{2001}$ & $\textbf{2199}$ & $\textbf{9899}$ & $\textbf{10001}$ & $\textbf{130192}$ & $\textbf{129998}$ & $\textbf{177189}$ & $\textbf{177190}$ \tabularnewline
\hline
\end{tabular}
\vspace{-0.1cm}
\caption{Incremental learning NSPDA (without hints) performance results. Each value is a measurement of the average number of characters required to reach convergence  (M1 = $2$nd Order NSPDA, M2 = $3$rd order NSDPA).}
\label{il_results}

\centering

\begin{tabular}{l|r|r||r|r||r|r||r|r}
\multicolumn{1}{l}{}&\multicolumn{2}{|c}{\begin{tabular}[x]{@{}c@{}}\textbf{$Palindrome$}\\\end{tabular}}&\multicolumn{2}{|c}{\begin{tabular}[x]{@{}c@{}}\textbf{$a^nb^n$}\\\end{tabular}}&\multicolumn{2}{|c}{\begin{tabular}[x]{@{}c@{}}\textbf{$a^nb^ncb^ma^m$}\\\end{tabular}}&\multicolumn{2}{|c}{\begin{tabular}[x]{@{}c@{}}\textbf{$a^{n+m}b^nc^m$}\\\end{tabular}}\tabularnewline
\multicolumn{1}{l}{\textbf{Regularization Method}}&\multicolumn{1}{|c}{\begin{tabular}[x]{@{}c@{}}\textbf{M1}\\\end{tabular}}&\multicolumn{1}{|c}{\begin{tabular}[x]{@{}c@{}}\textbf{M2}\\\end{tabular}}&\multicolumn{1}{|c}{\begin{tabular}[x]{@{}c@{}}\textbf{M1}\\\end{tabular}}&\multicolumn{1}{|c}{\begin{tabular}[x]{@{}c@{}}\textbf{M2}\\\end{tabular}}&\multicolumn{1}{|c}{\begin{tabular}[x]{@{}c@{}}\textbf{M1}\\\end{tabular}}&\multicolumn{1}{|c}{\begin{tabular}[x]{@{}c@{}}\textbf{M2}\\\end{tabular}}&\multicolumn{1}{|c}{\begin{tabular}[x]{@{}c@{}}\textbf{M1}\\\end{tabular}}&\multicolumn{1}{|c}{\begin{tabular}[x]{@{}c@{}}\textbf{M2}\\\end{tabular}}\tabularnewline
\hline
\textit{w/o reg} & $4.55$ & $2.99$ & $1.28$ & $1.55$ & $5.51$ & $4.19$ & $2.18$ & $2.00$ \tabularnewline
\textit{w reg} & $\textbf{0.00}$ & $\textbf{0.00}$ & $\textbf{0.06}$ & $\textbf{0.01}$ & $\textbf{0.99}$ & $\textbf{0.00}$ & $\textbf{0.09}$ & $\textbf{0.00}$ \tabularnewline
\hline
\end{tabular}
\vspace{-0.1cm}
\caption{Mean classification error for an NSPDA w/ \& w/o adaptive noise (tested on string length up to $T=60$).}
\label{noisereg_results}

\centering

\begin{tabular}{l|r|r||r|r||r|r||r|r||r|r}
\multicolumn{1}{l}{}&\multicolumn{2}{|c}{\begin{tabular}[x]{@{}c@{}}\textbf{$Palindrome$}\\\end{tabular}}&\multicolumn{2}{|c}{\begin{tabular}[x]{@{}c@{}}\textbf{$a^nb^n$}\\\end{tabular}}&\multicolumn{2}{|c}{\begin{tabular}[x]{@{}c@{}}\textbf{$a^nb^ncb^ma^m$}\\\end{tabular}}&\multicolumn{2}{|c}{\begin{tabular}[x]{@{}c@{}}\textbf{$a^{n+m}b^nc^m$}\\\end{tabular}} &\multicolumn{2}{|c}{\begin{tabular}[x]{@{}c@{}}\textbf{$Parenthesis$}\\\end{tabular}}\tabularnewline
\multicolumn{1}{l}{\textbf{RNN Type}}&\multicolumn{1}{|c}{\begin{tabular}[x]{@{}c@{}}\textbf{Train}\\\end{tabular}}&\multicolumn{1}{|c}{\begin{tabular}[x]{@{}c@{}}\textbf{Test}\\\end{tabular}}&\multicolumn{1}{|c}{\begin{tabular}[x]{@{}c@{}}\textbf{Train}\\\end{tabular}}&\multicolumn{1}{|c}{\begin{tabular}[x]{@{}c@{}}\textbf{Test}\\\end{tabular}}&\multicolumn{1}{|c}{\begin{tabular}[x]{@{}c@{}}\textbf{Train}\\\end{tabular}}&\multicolumn{1}{|c}{\begin{tabular}[x]{@{}c@{}}\textbf{Test}\\\end{tabular}}&\multicolumn{1}{|c}{\begin{tabular}[x]{@{}c@{}}\textbf{Train}\\\end{tabular}}&\multicolumn{1}{|c}{\begin{tabular}[x]{@{}c@{}}\textbf{Test}\\\end{tabular}}&\multicolumn{1}{|c}{\begin{tabular}[x]{@{}c@{}}\textbf{Train}\\\end{tabular}}&\multicolumn{1}{|c}{\begin{tabular}[x]{@{}c@{}}\textbf{Test}\\\end{tabular}}\tabularnewline
\hline
\textit{RNN} & $0.00$ & $78.2$ & $0.00$ & $74.11$ & $0.00$ & $83.33$ & $0.00$ & $73.69$ & $30.72$ & $99.96$\tabularnewline
\textit{LSTM} & $0.00$ & $12.58$ & $0.00$ & $13.26$ & $0.00$ & $14.22$ & $0.00$ & $10.56$ & $48.92$ & $97.88$\tabularnewline
\textit{LSTM-p} & $0.00$ & $8.69$ & $0.00$ & $11.25$ & $0.00$ & $13.99$ & $0.00$ & $12.88$ & $49.68$ & $99.00$\tabularnewline
\textit{GRU} & $0.00$ & $14.99$ & $0.00$ & $14.89$ & $0.00$ & $19.22$ & $0.00$ & $14.00$ & $43.21$ & $98.70$\tabularnewline
\textit{Stack RNN 40+10} & $0.00$ & $4.99$ & $0.00$ & $3.01$ & $0.00$ & $34.19$ & $0.00$ & $58.66$ & $10.25$ & $9.38$\tabularnewline
\textit{Stack RNN 40+10+ rounding} & $0.00$ & $0.09$ & $0.00$ & $0.89$ & $0.00$ & $1.01$ & $0.00$ & $0.79$ & $4.03$ & $3.968$\tabularnewline
\textit{listRNN 40+5} & $0.00$ & $0.39$ & $0.00$ & $2.29$ & $0.00$ & $19.63$ & $0.00$ & $1.27$ & $4.89$ & $7.45$\tabularnewline
\textit{$2$nd Order RNN} & $0.00$ & $9.26$ & $0.00$ & $8.51$ & $0.00$ & $17.52$ & $0.00$ & $11.17$ & $27.89$ & $37.42$\tabularnewline
\textit{$2$nd Order RNN reg \textbf{(ours)}} & $0.00$ & $1.88$ & $0.00$ & $2.09$ & $0.00$ & $2.19$ & $0.00$ & $0.99$ & $21.69$ & $27.59$\tabularnewline
\textit{NNPDA} & $0.00$ &  $7.00$ & $0.00$ &  $15.25$ & $0.00$ &  $17.49$ & $0.00$ &  $55.28$ & $5.96$ &  $29.21$ \tabularnewline
\textit{NNPDA} reg \textbf{(ours)} & $0.00$ &  $4.28$ & $0.00$ &  $14.20$ & $0.00$ &  $13.00$ & $0.00$ &  $41.01$ & $5.62$ &  $27.09$ \tabularnewline
\textit{NSPDA, M1 \textbf{(ours)}} & $0.00$ & $0.00$ & $0.00$ & $0.06$ & $0.00$ & $0.99$ & $0.00$ & $0.09$ & $0.58$ & $2.58$\tabularnewline
\textit{NSPDA, M2 \textbf{(ours)}} & $\textbf{0.00}$ & $\textbf{0.00}$ & $\textbf{0.00}$ & $\textbf{0.01}$ & $\textbf{0.00}$ & $\textbf{0.00}$ & $\textbf{0.00}$ & $\textbf{0.00}$ & $\textbf{0.01}$ & $\textbf{0.88}$\tabularnewline
\hline
\end{tabular}
\vspace{-0.1cm}
\caption{Mean classification error for various recurrent architectures when tested on strings of length up to $T=60$.}
\label{comp_results}
\end{table*}

\section{Experimental Details}
\label{sec:experiments}

\noindent
We focused on five context-free grammars, some labeled as Dyck(2) languages, which are some of the more difficult CFGs to recognize. For each grammatical inference task, we create a dataset that contains $1987$ positive and $2021$ negative (string) samples. Each sequence was of length $T$ which was sampled via $T \sim U(1,21)$, where $U(a,b)$ is the uniform distribution defined over the interval $[a,b]$. From the samples generated, we randomly sampled a subset from the total number of tokens generated.

\noindent

The number of state neurons for a second order NSPDA is set according to the following formula: $J = M + \sim U(12,29) $. For a third order NSDPA, the number of state neurons was set according to:  $J = M + \sim U(2,6) $. 

All models made use of the iterative refinement loss (Equation \ref{eqn:if_loss}, with $K=4$), weight updates were computed using whichever algorithm, i.e., BPTT, truncated BPTT (TBPTT) ($50$ steps back in time), RTRL, or UORO, yielded best performance for a given model. 
For higher order networks, UORO performed better and we use this to optimize all RNNs of this type in this study\footnote{For all first order RNNs, we found BPTT worked best and use that to train all RNNs of this type in our experiments.} (in the appendix, we offer a comparison of the various weight update rules when training an NSPDA). 
Gradients were hard clipped to $13$. 
Parameters were updated using stochastic gradient descent (SGD) which made use of the stochastic learning rate annealing scheme proposed in \cite{ororbia2019iterdecode} with initial learning rate of $0.1005000321$. 
All models were trained for a maximum of $500$ epochs (or until convergence was reached, which was marked as 100\% training accuracy). Experiments for each and every model was repeated $5$ times.

All of our models used our proposed rule encoding scheme and all of the RNNs were trained using our proposed two-stage incremental learning procedure. In Table \ref{il_results}, to demonstrate the value of our proposed two stage incremental training procedure (\emph{2-IL}), we compare an NSPDA trained without any incremental learning, one with ours, and one with the incremental learning approach (\emph{IL}) proposed in \cite{das1993using} and find that the our approach yields the best results across all grammars. 
All higher-order RNNs made use of our proposed adaptive noise regularizer, though in Table \ref{noisereg_results}, we examine how the NSPDA performs with and without the proposed regularizer. With respect to the hints used, for all tables presented in the main paper, whenever hint usage is indicated, we mean Hint \#2 (which worked the best empirically). 
In the appendix, we provide a detailed breakdown and ablation for all of the models investigated in this paper. Specifically, we present results for models that were trained with and without our regularizer as well as under various hint insertion conditions (no hints, Hint \#1, and Hint \#2).

\noindent
\textbf{Baseline Algorithms:} In order to provide the proper context do demonstrate the effectiveness of our proposed NSPDA, we conduct a thorough comparison of our model to as many baseline RNN models as possible. 
These models include a plethora of first order RNNs such as variations of the stack-RNN \cite{joulin2015inferring} (depth $k=2$, all other metaparameters set according to original source) 
including the two variant models as well as the linked-list model (using the same model labels as the original paper), the Long Short Term Memory RNN \cite{hochreiter1997long}  with (LSTM) and without peepholes (LSTM-p), the Gated Recurrent Unit (GRU) RNN \cite{chung2014empirical}, and a simple Elman RNN.
We also compared to gated first order RNNs with multiplicative units, but due to space constraints, we report these results in the appendix.
We furthermore compare against second order RNNs with ($2$nd Order RNN) and without regularization ($2$nd Order RNN reg), as well as the classical NNPDA with and without regularization (NNPDA reg).
All baselines RNNs had a single layer of $\leq 50$ neurons and individual hyperparameters for each was optimized based on validation set performance.

\section{Results and Discussion}
\label{sec:results}
To the best of our knowledge, we are the first to conduct a comparison across such a wide variety of RNN models of both first, second, and third order, with and without external (stack-based) memory.
For simple algorithmic patterns (non-Dyck(2) CFGs), first order RNNs like the LSTM and GRU perform reasonably well, primarily because they utilize dynamic counting \cite{lstmcfg,sennhauser2018evaluating} but yet do not learn any state transitions. This is evidenced when considering their performance on on the complex Dyck(2) CFG where the majority of RNNs exhibit great difficulty in generalizing to longer sequences.
These results do corroborate those of prior work, specifically those that demonstrate that the LSTM essentially performs a form of dynamic counting, making it ill-suited to recognizing complex grammars \cite{Lstmdynamiccounting}.


As pointed out by \cite{Lstmdynamiccounting} there is a strong need for neural architectures with external memory, i.e., a stack, to solve complex CFGs but, in this study, we furthermore argue that prior knowledge is also needed as well. This makes sense given that is known that prior information often leads to greatly improved reasoning and better generalization \cite{manning2019nsm}.
The stack and list RNNs do make use of (continuous) external memory (in fact, multiple stack/lists) but, theoretically, only one stack should be sufficient to recognize a PDA of any arbitrary length while a 2-stack PDA is as powerful as a Turning machine \cite{hopcroft2pda}.
However, quite surprisingly, \textbf{a stack-RNN with even 10 stacks has difficulty in generalizing to a complex grammar}. This lines up with the theory -- \cite{hopcroft2pda} has proven that adding any more than $2$ stacks to a PDA does not provide any further computational advantage.


Finally, it is impressive to see that high order RNNs coupled with external memory, particularly with a discrete stack structure (as opposed to a continuous stack like that of the stack-RNN), perform so well across all CFGs. It is important to note that even the way our state-based RNN operates is markedly different than the way those of the past did -- the NSPDA works as a next-step prediction model, which allows us to use the powerful iterative refinement procedure as a way to aggressively error correct its states when predicting string validity (at least during training time).
Table \ref{comp_results} shows that \textbf{our NSPDA model generalizes very well when trained on sequences of length $T \leq 21$ but tested on sequences on length up to $T=60$}.
Finally, our results demonstrate the value of rule insertion, which, as we see empirically, in some cases, improved convergence speed by a wide margin. 


\section{Conclusions}
\label{sec:conclusion}
In this work, we proposed the neural state pushdown automate (NSPDA) and its learning process, which utilizes an iterative refinement-based loss function, a two-stage incremental training procedure, an adaptive noise regularization scheme (which works with any higher order network), and a method for stably encoding rules into the model itself.

Our experimental results, which focused on context-free grammars (CFGs), demonstrate that prior knowledge is essential to learning memory-augmented that recognize complex CFGs well. Notably, we have empirically demonstrated the expressvity and flexibility of a high order temporal neural model that learns how to manipulate an external discrete stack.
While our proposed neural model works with a discrete stack, our model's underlying framework could be extended to manipulate other kinds of data structures, a subject of future work. When training on various CFGs, the state-based neural models we optimize converge faster and are more expressive than even powerful classical models such as the neural network pushdown automaton. Furthermore, we have shown that modern-day, popular recurrent network structures (all of which are first order) struggle greatly to recognize complex grammars.These discovered limitations of first order RNNs indicates that ANN research should consider the exploration of more expressive, memory-augmented models that offer ways to better integrate prior knowledge. 


\begin{small}
\fontsize{9.0pt}{10.5pt} \selectfont
\bibliographystyle{unsrt}
\bibliography{nspda_main.bib} 
\end{small}

\newpage

\section*{Appendix}

\section{Additional Results}
\label{sec:extra_results}
In Table \ref{comp_results2}, we report an expansion of the model performance table that appears in the main paper. In it, we report the performance of 3 modern gated RNNs with multiplicative gating units, i.e., MI-RNN, MI-LSTM, MI-GRU. Interestingly enough, one could consider the multiplicative units to be a crude approximation of second order state neurons.

Table \ref{pg_results} shows results for stably programming the weights of the NSPDA which, in effect, demonstrates that a programmed NSPDA (without learning) is equivalent to complex grammar PDA.

In the other table (Table \ref{lr_results}), we highlight how various learning algorithms affect the generalization ability of higher order recurrent networks. Here, we compare back-propagation through time (BPTT) to other online learning algorithms such as real time recurrent learning (RTRL) and unbiased online recurrent optimization (UORO). We describe these procedures in further detail in the next section.

Notably, in our experiments, we observed that UORO boosts performance for higher order recurrent networks, while being faster than RTRL, the original algorithm-of-choice when training higher order, state-based models.  Furthermore, we remark that truncated BPTT (TBPTT), for some CFGs, can actually slightly improve model performance over BPTT (but in ohers, such as is the case for the palindrome CFG, lead to worse generalization).
\begin{table*}[!t]
\centering
    \setlength{\extrarowheight}{2pt}
\begin{tabular}{l|r|r|r||r|r|r||r|r|r||r|r|r}
\multicolumn{1}{l}{}&\multicolumn{3}{|c}{\begin{tabular}[x]{@{}c@{}}\textbf{$Palindrome$}\\\end{tabular}}&\multicolumn{3}{|c}{\begin{tabular}[x]{@{}c@{}}\textbf{$a^nb^n$}\\\end{tabular}}&\multicolumn{3}{|c}{\begin{tabular}[x]{@{}c@{}}\textbf{$a^nb^ncb^ma^m$}\\\end{tabular}}&\multicolumn{3}{|c}{\begin{tabular}[x]{@{}c@{}}\textbf{$a^{n+m}b^nc^m$}\\\end{tabular}}\tabularnewline 
\hline
\multicolumn{1}{l}{\textbf{Model}}&\multicolumn{1}{|c}{\begin{tabular}[x]{@{}c@{}}\textbf{n=60}\\\end{tabular}}&\multicolumn{1}{|c}{\begin{tabular}[x]{@{}c@{}}\textbf{n=480}\\\end{tabular}}&\multicolumn{1}{|c}{\begin{tabular}[x]{@{}c@{}}\textbf{n=960}\\\end{tabular}}&\multicolumn{1}{|c}{\begin{tabular}[x]{@{}c@{}}\textbf{n=60}\\\end{tabular}}&\multicolumn{1}{|c}{\begin{tabular}[x]{@{}c@{}}\textbf{n=480}\\\end{tabular}}&\multicolumn{1}{|c}{\begin{tabular}[x]{@{}c@{}}\textbf{n=960}\\\end{tabular}}&\multicolumn{1}{|c}{\begin{tabular}[x]{@{}c@{}}\textbf{n=60}\\\end{tabular}}&\multicolumn{1}{|c}{\begin{tabular}[x]{@{}c@{}}\textbf{n=480}\\\end{tabular}}&\multicolumn{1}{|c}{\begin{tabular}[x]{@{}c@{}}\textbf{n=960}\\\end{tabular}}&\multicolumn{1}{|c}{\begin{tabular}[x]{@{}c@{}}\textbf{n=60}\\\end{tabular}}&\multicolumn{1}{|c}{\begin{tabular}[x]{@{}c@{}}\textbf{n=480}\\\end{tabular}}&\multicolumn{1}{|c}{\begin{tabular}[x]{@{}c@{}}\textbf{n=960}\\\end{tabular}}\tabularnewline 
\textit{$2$nd Order NSPDA} & $0.0$ & $0.0$ & $0.0$ & $0.0$ & $0.0$ & $0.0$ & $0.0$ & $0.0$ & $0.0$& $0.0$ & $0.0$ & $0.0$\tabularnewline 
\textit{$3$rd Order NSPDA} & $0.0$ & $0.0$ & $0.0$ & $0.0$ & $0.0$ & $0.0$ & $0.0$ & $0.0$ & $0.0$ & $0.0$ & $0.0$ & $0.0$ \tabularnewline 
\hline
\end{tabular}

\vspace{-0.1cm}
\caption{Mean classification error results when using a programmed NSPDA (lower is better).}
\label{pg_results}

\centering
\setlength{\extrarowheight}{2pt}
\begin{tabular}{l|r|r||r|r||r|r||r|r}
\multicolumn{1}{l}{}&\multicolumn{2}{|c}{\begin{tabular}[x]{@{}c@{}}\textbf{$Palindrome$}\\\end{tabular}}&\multicolumn{2}{|c}{\begin{tabular}[x]{@{}c@{}}\textbf{$a^nb^n$}\\\end{tabular}}&\multicolumn{2}{|c}{\begin{tabular}[x]{@{}c@{}}\textbf{$a^nb^ncb^ma^m$}\\\end{tabular}}&\multicolumn{2}{|c}{\begin{tabular}[x]{@{}c@{}}\textbf{$a^{n+m}b^nc^m$}\\\end{tabular}}\tabularnewline
\hline
\multicolumn{1}{l}{\textbf{Learning Algorithm}}&\multicolumn{1}{|c}{\begin{tabular}[x]{@{}c@{}}\textbf{M1}\\\end{tabular}}&\multicolumn{1}{|c}{\begin{tabular}[x]{@{}c@{}}\textbf{M2}\\\end{tabular}}&\multicolumn{1}{|c}{\begin{tabular}[x]{@{}c@{}}\textbf{M1}\\\end{tabular}}&\multicolumn{1}{|c}{\begin{tabular}[x]{@{}c@{}}\textbf{M2}\\\end{tabular}}&\multicolumn{1}{|c}{\begin{tabular}[x]{@{}c@{}}\textbf{M1}\\\end{tabular}}&\multicolumn{1}{|c}{\begin{tabular}[x]{@{}c@{}}\textbf{M2}\\\end{tabular}}&\multicolumn{1}{|c}{\begin{tabular}[x]{@{}c@{}}\textbf{M1}\\\end{tabular}}&\multicolumn{1}{|c}{\begin{tabular}[x]{@{}c@{}}\textbf{M2}\\\end{tabular}}\tabularnewline 
\hline
\textit{BPTT} & $2.02$ & $1.99$ & $2.23$ & $2.55$ & $2.99$ & $2.79$ & $2.99$ & $1.59$ \tabularnewline 
\textit{TBPTT} & $2.59$ & $2.81$ & $1.05$ & $1.29$ & $2.97$ & $2.02$ & $1.58$ & $1.11$ \tabularnewline 
\textit{RTRL} & $0.02$ & $0.19$ & $0.09$ & $0.10$ & $1.85$ & $0.07$ & $0.09$ & $0.01$ \tabularnewline 
\textit{UORO} & $0.00$ & $0.00$ & $0.06$ & $0.01$ & $0.99$ & $0.00$ & $0.09$ & $0.00$ \tabularnewline 
\hline
\end{tabular}
\vspace{-0.1cm}
\caption{Mean classification error for the NSPDA trained via various learning algorithms (tested on string length up to $T=60$).}
\label{lr_results}

\centering
\setlength{\extrarowheight}{2pt}
\begin{tabular}{l|r|r||r|r||r|r||r|r||r|r}
\multicolumn{1}{l}{}&\multicolumn{2}{|c}{\begin{tabular}[x]{@{}c@{}}\textbf{$Palindrome$}\\\end{tabular}}&\multicolumn{2}{|c}{\begin{tabular}[x]{@{}c@{}}\textbf{$a^nb^n$}\\\end{tabular}}&\multicolumn{2}{|c}{\begin{tabular}[x]{@{}c@{}}\textbf{$a^nb^ncb^ma^m$}\\\end{tabular}}&\multicolumn{2}{|c}{\begin{tabular}[x]{@{}c@{}}\textbf{$a^{n+m}b^nc^m$}\\\end{tabular}} &\multicolumn{2}{|c}{\begin{tabular}[x]{@{}c@{}}\textbf{$Parenthesis$}\\\end{tabular}}\tabularnewline
\multicolumn{1}{l}{\textbf{RNN Type}}&\multicolumn{1}{|c}{\begin{tabular}[x]{@{}c@{}}\textbf{Train}\\\end{tabular}}&\multicolumn{1}{|c}{\begin{tabular}[x]{@{}c@{}}\textbf{Test}\\\end{tabular}}&\multicolumn{1}{|c}{\begin{tabular}[x]{@{}c@{}}\textbf{Train}\\\end{tabular}}&\multicolumn{1}{|c}{\begin{tabular}[x]{@{}c@{}}\textbf{Test}\\\end{tabular}}&\multicolumn{1}{|c}{\begin{tabular}[x]{@{}c@{}}\textbf{Train}\\\end{tabular}}&\multicolumn{1}{|c}{\begin{tabular}[x]{@{}c@{}}\textbf{Test}\\\end{tabular}}&\multicolumn{1}{|c}{\begin{tabular}[x]{@{}c@{}}\textbf{Train}\\\end{tabular}}&\multicolumn{1}{|c}{\begin{tabular}[x]{@{}c@{}}\textbf{Test}\\\end{tabular}}&\multicolumn{1}{|c}{\begin{tabular}[x]{@{}c@{}}\textbf{Train}\\\end{tabular}}&\multicolumn{1}{|c}{\begin{tabular}[x]{@{}c@{}}\textbf{Test}\\\end{tabular}}\tabularnewline 
\hline
\textit{RNN} & $0.00$ & $78.2$ & $0.00$ & $74.11$ & $0.00$ & $83.33$ & $0.00$ & $73.69$ & $30.72$ & $99.96$\tabularnewline 
\textit{LSTM} & $0.00$ & $12.58$ & $0.00$ & $13.26$ & $0.00$ & $14.22$ & $0.00$ & $10.56$ & $48.92$ & $97.88$\tabularnewline 
\textit{LSTM-p} & $0.00$ & $8.69$ & $0.00$ & $11.25$ & $0.00$ & $13.99$ & $0.00$ & $12.88$ & $49.68$ & $99.00$\tabularnewline 
\textit{GRU} & $0.00$ & $14.99$ & $0.00$ & $14.89$ & $0.00$ & $19.22$ & $0.00$ & $14.00$ & $43.21$ & $98.70$\tabularnewline 
\textit{Stack RNN 40+10} & $0.00$ & $4.99$ & $0.00$ & $3.01$ & $0.00$ & $34.19$ & $0.00$ & $58.66$ & $10.25$ & $9.38$\tabularnewline 
\textit{Stack RNN 40+10+ rounding} & $0.00$ & $0.09$ & $0.00$ & $0.89$ & $0.00$ & $1.01$ & $0.00$ & $0.79$ & $4.03$ & $3.968$\tabularnewline 
\textit{listRNN 40+5} & $0.00$ & $0.39$ & $0.00$ & $2.29$ & $0.00$ & $19.63$ & $0.00$ & $1.27$ & $4.89$ & $7.45$\tabularnewline 
\textit{MI-RNN} & $0.00$ & $75.69$ & $0.00$ & $70.26$ & $0.00$ & $76.69$ & $0.00$ & $73.01$ & $29.58$ & $99.92$\tabularnewline 
\textit{MI-LSTM} & $0.00$ & $9.99$ & $0.00$ & $10.86$ & $0.00$ & $13.55$ & $0.00$ & $14.22$ & $47.83$ & $99.80$\tabularnewline 
\textit{MI-GRU} & $0.00$ & $16.22$ & $0.00$ & $13.29$ & $0.00$ & $20.02$ & $0.00$ & $14.83$ & $42.88$ & $99.20$\tabularnewline 
\textit{$2$nd Order RNN} & $0.00$ & $9.26$ & $0.00$ & $8.51$ & $0.00$ & $17.52$ & $0.00$ & $11.17$ & $27.89$ & $37.42$\tabularnewline 
\textit{$2$nd Order RNN reg \textbf{(ours)}} & $0.00$ & $1.88$ & $0.00$ & $2.09$ & $0.00$ & $2.19$ & $0.00$ & $0.99$ & $21.69$ & $27.59$\tabularnewline 
\textit{NNPDA} & $0.00$ &  $7.00$ & $0.00$ &  $15.25$ & $0.00$ &  $17.49$ & $0.00$ &  $55.28$ & $5.96$ &  $29.21$ \tabularnewline 
\textit{NNPDA} reg \textbf{(ours)} & $0.00$ &  $4.28$ & $0.00$ &  $14.20$ & $0.00$ &  $13.00$ & $0.00$ &  $41.01$ & $5.62$ &  $27.09$ \tabularnewline 
\textit{NSPDA, M1 \textbf{(ours)}} & $0.00$ & $0.00$ & $0.00$ & $0.06$ & $0.00$ & $0.99$ & $0.00$ & $0.09$ & $0.58$ & $2.58$\tabularnewline 
\textit{NSPDA, M2 \textbf{(ours)}} & $\textbf{0.00}$ & $\textbf{0.00}$ & $\textbf{0.00}$ & $\textbf{0.01}$ & $\textbf{0.00}$ & $\textbf{0.00}$ & $\textbf{0.00}$ & $\textbf{0.00}$ & $\textbf{0.01}$ & $\textbf{0.88}$\tabularnewline
\hline 
\end{tabular}
\caption{Mean classification error for various recurrent architectures when tested on strings of length up to $T=60$.}
\label{comp_results2}
\end{table*}
\section{On Training Algorithms}
\label{sec:training_algos}
For all of the RNNs we study, we compared their (validation) performance when using various online and offline based learning algorithms. As mentioned in the last section, we found that UORO worked best for the NSPDA, which is advantageous in that UORO is faster than RTRL (even largely in terms of complexity) and does not require model unfolding like the popular and standard BPTT/TBPTT algorithms do. These results, again, are summarized in Table \ref{lr_results}.

Below we briefly describe the non-standard approaches to learning RNNs, specifically RTRL and UORO. Notably, we are the first to implement and adapt UORO in calculating the updates to the weights of higher order networks.

\subsection{Real-Time Recurrent Learning}
\label{sec:rtrl}
Real-time recurrent learning (RTRL) is a classical online learning procedure for training RNNs \cite{williams1989rtrl}. The aim is to optimize the parameters $\Theta$ of a state-based model in order to minimize a total (sequence) loss. The state model is abstract to the following function:
\begin{align}
\mathbf{z}_{t+1} = F_{state}(\mathbf{x}_{t+1},\mathbf{z}_{t},\Theta) \mbox{.} \label{rtrl:eqn1}
\end{align}
RTRL computes the derivative of the model's states and outputs with respect to the synaptic weights during the model's forward computation, as data points in the sequence are processed iteratively, i.e., without any unfolding as in BPTT.
When the task is next step prediction (predict $\mathbf{x}_t$ given a history $\mathbf{x}_{<t}$), the loss $L$ to optimize, using RTRL, is defined as follows: 
\begin{multline}
\frac { \partial L _ { t + 1 } } { \partial \Theta } = \frac { \partial L_{t + 1}(\mathbf{y}_{t + 1}, \mathbf{y}_{t + 1}^{*} )} {\partial \mathbf{y}}  \otimes \bigg( \frac{\partial F_{\text{out}}(\mathbf{x}_{t + 1}, \mathbf{z}_{t}, \Theta)} {\partial \mathbf{z}_t} \frac{ \partial \mathbf{z}_{t} } {\partial \Theta} + \frac {\partial F_ {\text{out}}(\mathbf{x}_{t + 1} , \mathbf{z}_{t} , \Theta)}{\partial \Theta} \bigg) \mbox{.} \label{rtrl:eqn2}
\end{multline}
Once we differentiate Equation \ref{rtrl:eqn1} with respect to $\Theta$, we obtain:
\begin{align}
\frac{\partial \mathbf{z}_t+1}{\partial \Theta} = \frac{\partial F_{\text {state}}(\mathbf{x}_{t+1}, \mathbf{z}_{t}, \Theta)}{\partial \Theta} + \frac{\partial F_{\text {state}}(\mathbf{x}_{t+1}, \mathbf{z}_{t}, \Theta)}{\partial \mathbf{z}_t} \otimes \frac{\partial \mathbf{z}_t}{\partial \Theta} \mbox{.} \label{rtrl:eqn3}
\end{align}
Where at each time we compute $\frac{\partial \mathbf{z}_t}{\partial \Theta}$ based on $\frac{\partial \mathbf{z}_t-1}{\partial \Theta}$. These values are then used to directly compute $\frac{\partial \mathbf{z}_t+1}{\partial \Theta}$. 

The above is, in short, how RTRL calculates its gradients without resorting to backward transfer or computation graph unfolding (as in reverse-mode differentiation). Since the shape of $\frac{\partial \mathbf{z}_t}{\partial \Theta}$ is the same as $|z| \times |\Theta|$, for standard RNNs with $n$ hidden units, this calculation scales as ${n}^4$ (time complexity \cite{williams1995gradient}). This high complexity makes RTRL highly impractical for training very wide and very deep recurrent models.
However, in the case of a third order model like NSPDA (or an NNPDA), the number of states need for learning a target grammar are generally far fewer than those required of second or first order models (as we mentioned in the main paper). This means that a procedure such as RTRL is still applicable and useful at least for training RNNs to recognize context free grammars (of low input dimensionality).

\subsection{Unbiased Online Recurrent Optimization}
\label{sec:uoro}
Unbiased Online Recurrent Optimization (UORO) \cite{tallec2017uoro} uses a rank-one trick to approximate the operations need to make RTRL's gradient computation work. This trick helps to reduce the overall complexity of the at the price of increasing variance of its gradient estimates. 

When designing an optimizer like UORO, we start from the idea that for any given unbiased estimation of $\frac{\partial \mathbf{z}_t}{\partial \Theta}$, we can form a stochastic matrix $\tilde{Z}_t$ such that $\mathbb{E}(\tilde{Z}_{t}) = \frac{\partial \mathbf{z}_t}{\partial \Theta}$. 
Since Equation \ref{rtrl:eqn2} and \ref{rtrl:eqn3} are affine in $\frac{\partial \mathbf{z}_t}{\partial \theta}$, the ``unbiasedness'' (of gradient estimates) is preserved due to the linearlity of the expectation. 
Next, we compute the value of $\tilde{Z}_t$ and plug it into \ref{rtrl:eqn2} and \ref{rtrl:eqn3} to calculate the value for $\frac{\partial \mathbf{L}_t+1}{\partial \Theta}$ and $\frac{\partial \mathbf{z}_t+1}{\partial \Theta}$.
In a rank-one, unbiased approximation, at time step $t$, $\tilde{Z}_t = \tilde{z}_t \otimes \tilde{\Theta}_t$. To calculate $\hat{Z}_t+1$ at $t+1$, we plug in $\tilde{Z}_t$ into \ref{rtrl:eqn3}. Nonetheless, mathematically, the above is still not yet a rank-one approximation of RTRL. 

In order to finally obtain a proper rank-one approximation, one must use an additional, efficient approximation technique, proposed in \cite{ollivier2015training}, to rewrite the above equation as:
\begin{multline}
\tilde{Z}_{t+1} = \bigg( \rho_{0}\frac{\partial F_{\text{state}}(\mathbf{x}_{t+1},\mathbf{z}_{t},\theta)}{\partial \mathbf{z}} \tilde{\mathbf{z}}_t + \rho_{1}\nu \bigg) \otimes \bigg( \frac{\tilde{\theta}_t}{\rho_{0}} + \frac{(\nu)^T }{\rho_{1}}\frac{\partial F_{\text{state}}(\mathbf{x}_{t+1},\mathbf{z}_{t},\theta)}{\partial \theta} \bigg) \mbox{.}
\end{multline}
Note that $\nu$ is a vector of independent, random signs and $\rho$ contains $k$ positive numbers. Thus, the rank-one trick can be applied for any $\rho$.  In UORO, $\rho_0$ and $\rho_1$ are factors meant to control the variance of the estimator's computed approximate derivatives. In practice, we define $\rho_0$ as:
\begin{align}
\rho_{0} = \sqrt[]{\frac{\| \tilde{\theta}_t \|}{\| \frac{\partial F_{state}(\mathbf{x}_{t+1},\mathbf{z}_{t},\theta)}{\partial \mathbf{z}}\tilde{\mathbf{z}}\|}}
\end{align}
and $\rho_1$ is defined to be:
\begin{align}
\rho_{1} = \sqrt[]{\frac{\| (\nu)^T \frac{\partial F_{state}(\mathbf{x}_{t+1},\mathbf{z}_{t},\theta)}{\theta} \|}{\|\nu \|}}\mbox{.}
\end{align}
Initially, $\tilde{\mathbf{z}}_0 = 0$ and $\tilde{\Theta}_0 = 0$, which yields unbiased estimates at time $t = 0$. Given the construction of the UORO procedure, by induction, all subsequent estimates can be shown to be unbiased as well.

\end{document}